\documentclass[a4paper,fleqn]{cas-dc}

\usepackage[authoryear,longnamesfirst]{natbib}
\usepackage{booktabs}
\usepackage{siunitx}
\usepackage{float}
\usepackage{fancyhdr}
\fancypagestyle{first}{
  \fancyhf{}
  \fancyfoot[C]{\thepage}

}

\begin{document}
\let\WriteBookmarks\relax

\shorttitle{One-step-ahead forecasting of NEPSE Index using XGBoost}

\shortauthors{Malla et al.}

\title[mode=title]{XGBoost Forecasting of NEPSE Index Log Returns with Walk Forward Validation}

\tnotetext[1]{The authors declare that no funding was received for this research.}

\author[1]{Sahaj Raj Malla}[orcid=0000-0001-5980-6066]
\cormark[1]
\ead{sm03200822@student.ku.edu.np}
\credit{Conceptualization, Methodology, Software, Formal analysis, Investigation, Visualization, Writing -- original draft}

\author[1]{Shreeyash Kayastha}
\credit{Methodology, Software, Validation, Formal analysis, Writing -- review}

\author[1]{Rumi Suwal}
\credit{Data curation, Investigation, Writing -- review \& editing}

\author[1]{Harish Chandra Bhandari}
\credit{Supervision, Validation, Writing -- review \& editing}

\author[2]{Rajendra Adhikari}
\credit{Supervision, Resources, Writing -- review \& editing}

\affiliation[1]{%
    organization={Department of Mathematics, Kathmandu University},%
    addressline={P.O. Box 6250},%
    city={Dhulikhel},%
    postcode={45200},%
    state={Kavre, Bagmati Province},%
    country={Nepal}%
}

\affiliation[2]{%
    organization={Department of Physics, Kathmandu University},%
    addressline={P.O. Box 6250},%
    city={Dhulikhel},%
    postcode={45200},%
    state={Kavre, Bagmati Province},%
    country={Nepal}%
}

\cortext[cor1]{Corresponding author}

\begin{abstract}
This study develops a robust machine learning framework for one-step-ahead forecasting of daily log-returns in the Nepal Stock Exchange (NEPSE) Index using the XGBoost regressor. A comprehensive feature set is engineered, including lagged log-returns (up to 30 days) and established technical indicators such as short- and medium-term rolling volatility measures and the 14-period Relative Strength Index. Hyperparameter optimization is performed using Optuna with time-series cross-validation on the initial training segment. Out-of-sample performance is rigorously assessed via walk-forward validation under both expanding and fixed-length rolling window schemes across multiple lag configurations, simulating real-world deployment and avoiding lookahead bias. Predictive accuracy is evaluated using root mean squared error, mean absolute error, coefficient of determination ($R^{2}$), and directional accuracy on both log-returns and reconstructed closing prices. Empirical results show that the optimal configuration—an expanding window with 20 lags—outperforms tuned ARIMA and Ridge regression benchmarks, achieving the lowest log-return RMSE (0.013450) and MAE (0.009814) alongside a directional accuracy of 65.15\%. While the $R^{2}$ remains modest, consistent with the noisy nature of financial returns, primary emphasis is placed on relative error reduction and directional prediction. Feature importance analysis and visual inspection further enhance interpretability. These findings demonstrate the effectiveness of gradient boosting ensembles in modeling nonlinear dynamics in volatile emerging market time series and establish a reproducible benchmark for NEPSE Index forecasting.
\end{abstract}


\maketitle
\pagestyle{fancy}
\thispagestyle{first}

\section{Introduction}
\label{sec:introduction}
The Nepal Stock Exchange (NEPSE) Index serves as the principal benchmark for the equity market in Nepal, an emerging economy marked by relatively low market capitalization, limited liquidity, and heightened volatility compared to developed markets. These characteristics—including volatility clustering, persistence in conditional variance, and sensitivity to domestic political and macroeconomic factors—render accurate forecasting of index movements particularly challenging but essential for investors, policymakers, and financial institutions \citep{gajurel2021volatility}.

Financial time series forecasting has long been a focal point of econometric and machine learning research. Traditional statistical approaches, such as autoregressive integrated moving average (ARIMA) models, have been applied to NEPSE data with moderate success in capturing linear patterns, achieving directional accuracies around 50-60\% \citep{paudel2024predicting}. However, the inherent nonlinearity, non-stationarity, and noise in stock returns often limit their predictive performance. In recent years, machine learning techniques—particularly ensemble methods and deep learning architectures—have demonstrated superior ability to model complex relationships in financial data \citep{sezer2019financialtimeseriesforecasting}.

Prior studies on NEPSE forecasting have predominantly employed deep learning models, including Long Short-Term Memory (LSTM), Gated Recurrent Units (GRU), and Convolutional Neural Networks (CNN), for predicting closing prices or directional movements \citep{pokhrel2022predicting,shahi2020stock}; however, their reliance on large data volumes and limited interpretability makes them less suitable for low-liquidity emerging markets. Classification-based approaches incorporating news sentiment have also utilized algorithms such as support vector machines (SVM), random forests, and Extreme Gradient Boosting (XGBoost) to forecast index direction \citep{dahal2024predicting,zhao2021nepal}. Nonetheless, regression-based forecasting of continuous log-returns using tree-boosting methods remains underexplored in this context, despite XGBoost's success in financial applications such as stock return prediction in emerging markets \citep{ni2019xgboost,jing2021hybrid,jabeur2024forecasting}.

This study addresses this gap by applying XGBoost, a scalable gradient boosting algorithm known for its robustness in handling tabular time series data \citep{chen2016xgboost}, to one-step-ahead prediction of NEPSE Index daily log-returns. Log-returns are preferred over direct price-level forecasting due to their desirable statistical properties, including approximate normality and time-additivity, which facilitate modeling and mitigate issues with nonstationarity in financial time series. Features are engineered from lagged log-returns and established technical indicators, including rolling volatility measures and the Relative Strength Index (RSI). Hyperparameter optimization is performed using Optuna with time-series cross-validation on an initial training segment. Out-of-sample evaluation employs rigorous walk-forward validation \citep{bergmeir2018note}, simulating real-world deployment under both expanding and rolling window schemes, with multiple lag configurations examined for robustness.
The primary contributions of this study are as follows:
\begin{enumerate}
  \item A reproducible, feature-engineered XGBoost framework specifically tailored to forecasting an emerging market index, such as the NEPSE Index.
  \item A comprehensive out-of-sample assessment that incorporates regression metrics, directional accuracy, and reconstructed closing price levels.
  \item Valuable insights into feature importance and model interpretability within a volatile, low-liquidity market setting.
\end{enumerate}

The remainder of the paper is structured as follows. Section~\ref{sec:literature} reviews relevant literature. Section~\ref{sec:data} describes the dataset and feature engineering process. Section~\ref{sec:methodology} outlines the modeling framework, including hyperparameter tuning and walk-forward validation. Section~\ref{sec:results} presents empirical findings. Section~\ref{sec:discussion} interprets the results and implications, and Section~\ref{sec:conclusion} concludes.


\section{Literature Review}
\label{sec:literature}

Financial time series forecasting has evolved significantly, transitioning from traditional econometric models to advanced machine learning and deep learning techniques, driven by the need to capture nonlinearity, volatility clustering, and complex patterns in asset returns \citep{zhang2024deep}.

Early studies on the NEPSE Index primarily employed statistical models to characterize volatility and predictability. GARCH-family models have been widely applied to model conditional heteroskedasticity in NEPSE returns, confirming the presence of volatility clustering, persistence, and leverage effects \citep{gajurel2021volatility}. For instance, symmetric GARCH specifications performed well over extended periods, while asymmetric variants better captured pre- and post-event shocks, such as those following the 2015 earthquake \citep{gajurel2021volatility}. ARIMA-based approaches have also been utilized for level forecasting, with hybrid ARIMA-GARCH models demonstrating improved handling of both trend and volatility components \citep{paudel2024predicting}.

The advent of machine learning has shifted focus toward more flexible algorithms capable of modeling nonlinear relationships. Support vector regression (SVR) and artificial neural networks were among the earliest applications to NEPSE data, achieving reasonable predictive accuracy for sector-level prices \citep{pun2018nepal}. Subsequent research increasingly incorporated deep learning architectures, particularly recurrent neural networks suited to sequential data. LSTM, GRU, and CNN models have been compared for one-day-ahead closing price prediction, with LSTM variants often outperforming baselines in root mean squared error and directional accuracy \citep{pokhrel2022predicting,shahi2020stock}. While these deep learning approaches excel in capturing temporal dependencies, their reliance on large data volumes and limited interpretability renders them less ideal for low-liquidity emerging markets like NEPSE.

More recent contributions have explored hybrid and ensemble approaches. News headline sentiment has been integrated with classification algorithms—including logistic regression, SVM, random forests, and XGBoost—to forecast directional movements \citep{dahal2024predicting}. Similarly, \cite{kafle2024evaluating} provided an initial evaluation of XGBoost for direct Nepali stock price prediction. Nonetheless, regression-based forecasting of continuous log-returns using tree-boosting methods remains underexplored in this context, particularly under rigorous validation schemes.

In broader financial forecasting, gradient boosting methods such as XGBoost have gained prominence for tabular time series data, offering robustness, interpretability via feature importance, and strong performance in emerging markets \citep{chen2016xgboost,wang2020forecasting,nobre2019xgboost}. Systematic reviews highlight the superiority of deep learning over traditional methods post-2020, though ensemble and hybrid models continue to advance accuracy \citep{zhang2024deep,kumbure2022machine}.

A critical aspect of rigorous evaluation in financial forecasting is the avoidance of lookahead bias through proper validation schemes. Walk-forward (or rolling-origin) validation, which iteratively retrains models on expanding or rolling windows while testing on subsequent periods, provides a realistic assessment of out-of-sample performance and is recommended for nonstationary time series \citep{bergmeir2018note,hyndman2018forecasting}.

Despite progress, applications of gradient boosting ensembles to the NEPSE Index often lack strict rolling-origin evaluation. For instance, prior studies have frequently employed standard train-test splits without addressing the temporal evolution of model parameters. In contrast, the present study targets log-returns—preferred for their stationarity and time-additivity properties—with a gradient boosting approach (XGBoost) that offers superior interpretability through feature importance and robustness in low-liquidity settings. The adoption of rigorous walk-forward validation addresses prevalent methodological limitations in financial machine learning, ensuring that reported performance reflects realistic deployment scenarios rather than overstated in-sample results \citep{hyndman2018forecasting}.


\section{Data Description and Feature Engineering}
\label{sec:data}

The dataset comprises daily historical records of the NEPSE Index, sourced from the third-party portal NepseAlpha,\footnote{\url{https://nepsealpha.com/nepse-data}} which aggregates and provides export functionality for publicly available NEPSE market data. The records were compiled to form a comprehensive series spanning from July 20, 1997, to November 11, 2025, resulting in 6,393 observations after excluding non-trading days. Each record includes the trading date, opening price, high price, low price, adjusted closing price, percentage change, and trading volume.

As NepseAlpha aggregates publicly available official NEPSE data, this study utilizes the dataset solely for academic research.

For the forecasting task, only the date and closing price are retained from the raw data. Closing prices are converted to numeric values, and dates are converted to datetime format. The target variable is the daily log-return, calculated as
\begin{equation}
r_t = \log\left(\frac{P_t}{P_{t-1}}\right),
\end{equation}
where $P_t$ represents the closing price on day $t$. Log-returns are employed due to their desirable statistical properties, including approximate normality and time-additivity, which facilitate modeling in financial time series analysis \citep{tsay2005analysis}.

Feature engineering incorporates lagged values of the target variable and established technical indicators to capture potential autocorrelation, momentum, and volatility dynamics. Lagged log-returns are generated as
\begin{equation}
\text{lag}_k = r_{t-k}, \quad k = 1, \dots, L,
\end{equation}
with $L \in \{10, 20, 30\}$ varied across experimental configurations to empirically assess robustness in capturing short- to medium-term dependencies typical in daily equity returns.

The following technical indicators are computed:
\begin{itemize}
  \item Short-term volatility: rolling standard deviation of log-returns over 5 days (\verb|vol_5|).
  \item Medium-term volatility: rolling standard deviation of log-returns over 20 days (\verb|vol_20|).
  \item RSI over 14 periods:
  \begin{equation}
  \text{RSI}_{14} = 100 - \frac{100}{1 + \text{RS}}, \quad \text{RS} = \frac{\text{average gain over 14 days}}{\text{average loss over 14 days} + \epsilon},
  \end{equation}
  where $\epsilon = 10^{-8}$ prevents division by zero.
  \item Rolling mean of log-returns over 10 days (\verb|mean_10|).
\end{itemize}

All rolling features are computed using only past information and aligned to ensure no future data leakage. Features were checked for redundancy to ensure model efficiency. Any infinite or undefined values are replaced with NaN, and observations with missing values—arising primarily from the initial periods required for lags and rolling windows—are removed.

The temporal evolution of the NEPSE closing price series is illustrated in Figure~\ref{fig:nepse_price}, highlighting characteristic phases of growth, correction, and volatility typical of emerging markets.

The final dataset is split chronologically, reserving the last 20\% of observations exclusively for out-of-sample testing to avoid data leakage and ensure realistic forecasting conditions.


\section{Methodology}
\label{sec:methodology}

This section details the modeling framework employed for one-step-ahead forecasting of NEPSE Index log-returns. The primary model is the XGBoost regressor \citep{chen2016xgboost}, selected for its established efficacy in tabular time-series regression tasks. For benchmarking, tuned ARIMA models, Ridge regression, and a suite of deep learning architectures (CNN, LSTM, N-BEATS, TFT) are implemented as baselines, ensuring comparable evaluation protocols. All experiments are fully reproducible, with complete source code, processed datasets, models, and results publicly available.\footnote{\url{https://github.com/sahajrajmalla/nepse-xgboost-forecasting}}

\subsection{XGBoost Model and Hyperparameter Optimization}

The XGBoost regressor implements gradient tree boosting with regularization to mitigate overfitting. The objective function explicitly minimizes mean squared error (MSE) which is suitable for continuous log-return prediction.

Hyperparameter optimization is conducted using Optuna \citep{akiba2019optuna}, a Bayesian optimization framework with Tree-structured Parzen Estimator (TPE) sampling. Tuning is performed exclusively on the initial training segment (80\% of the dataset) to prevent data leakage. Cross-validation employs time-series split with 5 folds, respecting temporal order. A fixed random seed is used to ensure reproducibility.

The search space includes:
\begin{itemize}
  \item Number of estimators: 200--1200
  \item Maximum depth: 3--10
  \item Learning rate: 0.01--0.3 (log scale)
  \item Subsample and column sample by tree: 0.6--1.0
  \item Gamma: 0.0--5.0
  \item Minimum child weight: 1--10
  \item L1 regularization (\verb|reg_alpha|): 0.0--1.0
  \item L2 regularization (\verb|reg_lambda|): 0.0--2.0
\end{itemize}

Optimization runs for up to 60 trials. Fixed parameters include the squared error objective, histogram tree method, and full CPU utilization.

\subsection{Deep Learning Benchmarks}

To evaluate the predictive capability of the proposed XGBoost framework against modern neural architectures, four deep learning models were implemented: CNN, LSTM networks, Neural Basis Expansion Analysis for Time Series (N-BEATS), and Temporal Fusion Transformers (TFT). These architectures are widely utilized in financial forecasting for their ability to model complex temporal dependencies.

Consistent with the XGBoost approach, hyperparameter optimization for all deep learning models was performed using Optuna on the initial training segment. The search spaces included architecture-specific parameters (e.g., number of filters and kernel sizes for CNN, hidden units for LSTM, attention heads for TFT) as well as training parameters (learning rate, dropout, and batch size). To ensure strict comparability, these models were trained to forecast one-step-ahead log-returns and were subjected to the identical out-of-sample validation scheme described below. All deep learning experimental runs were executed on a high-performance workstation equipped with an NVIDIA Quadro RTX 6000 GPU.

\subsection{Walk-Forward Validation Framework}

Out-of-sample performance is assessed via walk-forward validation, a rolling-origin approach that simulates real-time deployment and avoids lookahead bias \citep{hyndman2018forecasting}. The dataset is split chronologically, reserving the final 20\% for testing.

For each one-step-ahead prediction in the test period:
\begin{itemize}
  \item \textbf{Expanding window}: The training set incorporates all observations up to the current step.
  \item \textbf{Rolling window}: The training set maintains a fixed length of 800 observations (approximately 3--4 years of daily data), sliding forward to balance adaptation to recent market conditions with sufficient training data volume.
\end{itemize}

The model is refitted at each step using the optimized hyperparameters from the initial tuning. Predicted log-returns are exponentiated with the prior closing price to reconstruct index levels:
\begin{equation}
\hat{P}_t = P_{t-1} \exp(\hat{r}_t),
\end{equation}
where \(\hat{r}_t\) is the predicted log-return. Price-level metrics are provided for interpretability but are secondary, as errors may accumulate in reconstructions; primary focus remains on log-return forecasts. This procedure is repeated across lag configurations (10, 20, 30 days) to evaluate robustness.

Benchmark models follow analogous protocols: ARIMA orders (p, q $\leq$ 5) are tuned via AIC minimization on the initial training segment, with returns confirmed stationary via Augmented Dickey-Fuller test (p-value < 0.01), requiring no differencing and no seasonal components due to daily frequency. Ridge regression regularization strength is optimized similarly to the other models.

\subsection{Performance Evaluation Metrics}

Forecast accuracy is quantified on both log-returns and reconstructed prices using:
\begin{itemize}
  \item Root Mean Squared Error (RMSE): \(\sqrt{\frac{1}{n} \sum (y_i - \hat{y}_i)^2}\)
  \item Mean Absolute Error (MAE): \(\frac{1}{n} \sum |y_i - \hat{y}_i|\)
  \item Coefficient of determination (R²): \(1 - \frac{\sum (y_i - \hat{y}_i)^2}{\sum (y_i - \bar{y})^2}\)
  \item Directional Accuracy: Percentage of correct sign predictions for log-returns, computed as
  \[
  \frac{1}{n} \sum_{i=1}^{n} I\!\left( \operatorname{sign}(r_i) = \operatorname{sign}(\hat{r}_i) \right),
  \]
  where \(I\) is the indicator function and \(r_i\) denotes the actual log-return.
\end{itemize}

Directional accuracy is computed directly on log-returns for primary evaluation, with equivalent results on price changes due to the monotonicity of the exponential transform. This metric is particularly relevant for financial applications, as it assesses the model's ability to predict market movement direction.

\section{Empirical Results}
\label{sec:results}

This section presents the out-of-sample performance of the proposed XGBoost framework for one-step-ahead forecasting of NEPSE Index log-returns, evaluated through rigorous walk-forward validation on the final 20\% of observations (approximately mid-2022 to November 2025). Comparisons are provided with tuned Ridge regression and ARIMA benchmarks.

\subsection{Data Characteristics}

\begin{figure*}[pos=t]
\centering
\includegraphics[width=\textwidth]{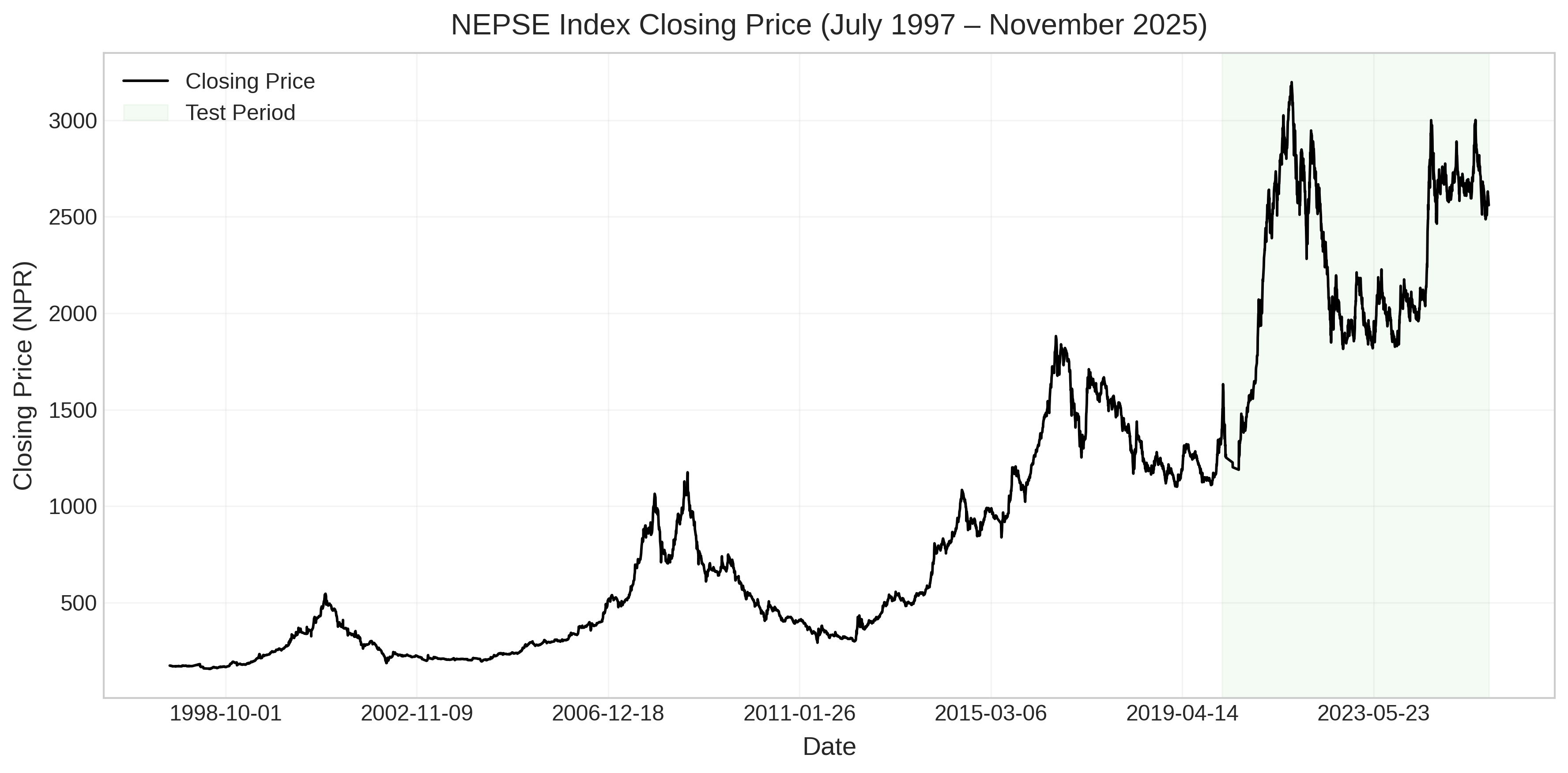}
\caption{Long-term evolution of the NEPSE Index closing price (July 1997--November 2025), with the out-of-sample test period lightly shaded.}
\label{fig:nepse_price}
\end{figure*}

Figure~\ref{fig:nepse_price} depicts the historical trajectory of the NEPSE Index closing price from July 1997 to November 2025. The series displays typical emerging market behavior: extended phases of upward growth, abrupt corrections, and pronounced volatility clusters, particularly in recent years.

\begin{figure}[pos=htbp]
\centering
\includegraphics[width=0.8\columnwidth]{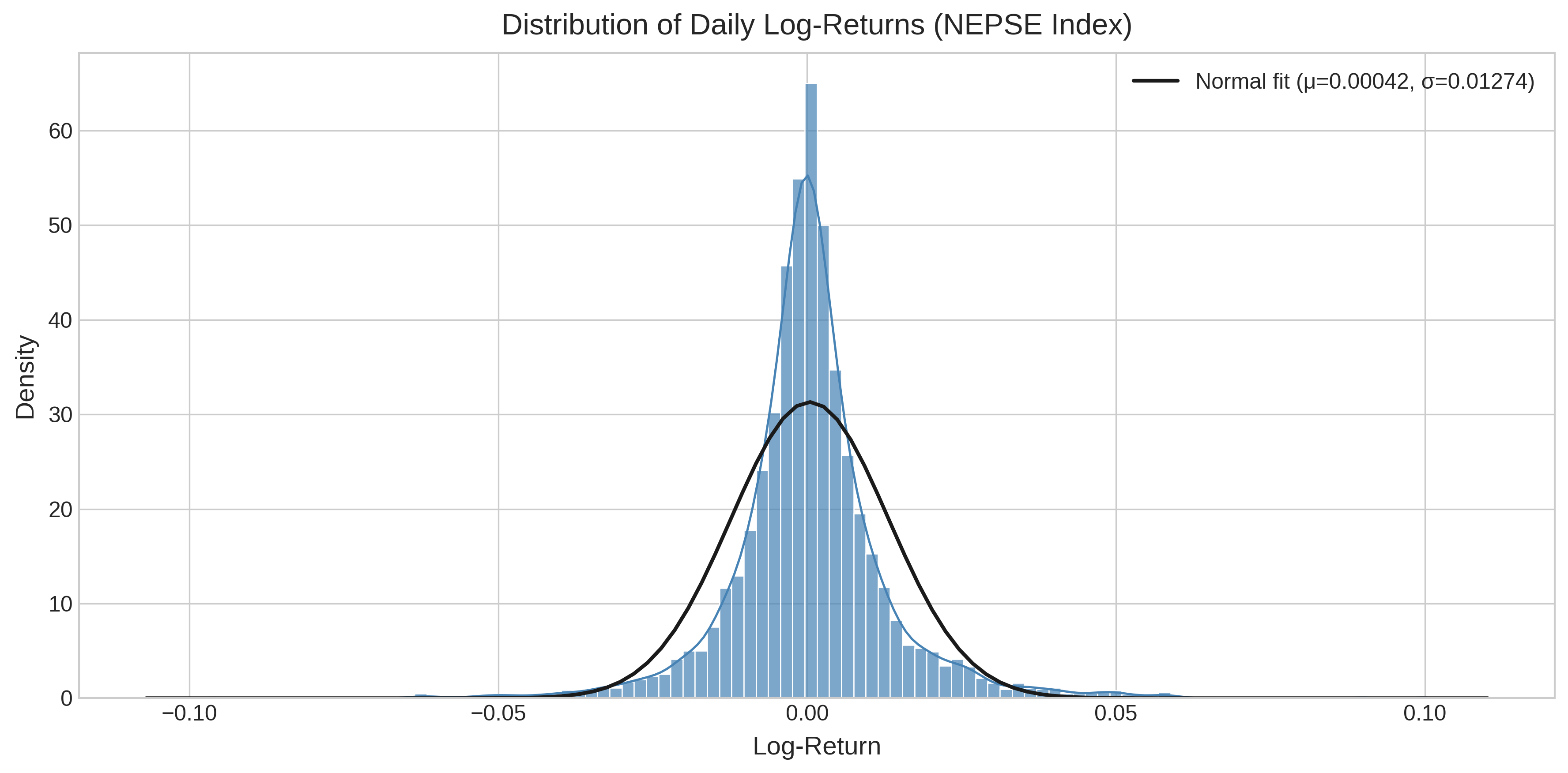}
\caption{Distribution of daily log-returns for the NEPSE Index, overlaid with a normal density fit.}
\label{fig:log_returns}
\end{figure}

Figure~\ref{fig:log_returns} shows the distribution of daily log-returns, which is nearly symmetric around a mean of approximately 0.00042 with a standard deviation of 0.01274.

\subsection{Hyperparameter Tuning Outcomes}

Hyperparameter optimization via Optuna consistently identified effective configurations across models. For the optimal XGBoost setup (expanding window, 20 lags), key parameters included a low learning rate (0.010), moderate tree depth (typically 7 to 9), and high subsampling ratios (approximately 0.95), with mild L1 and L2 regularization. These choices promote gradual learning and robustness in noisy financial data.

Ridge regression tuning yielded small to moderate regularization strengths (alpha values), indicating limited overfitting even with lagged features. ARIMA optimization converged to low-order specifications (predominantly AR(1) or AR(2) with d=0), consistent with the near-random-walk nature of log-returns and confirming stationarity (Augmented Dickey-Fuller p-value < 0.01).

\subsection{Out-of-Sample Performance on Log-Returns}

Table~\ref{tab:performance_log_returns} reports primary performance metrics on log-returns.

\begin{table*}[pos=htbp]
\centering
\caption{Out-of-Sample Performance on Log-Returns Across Models and Configurations}
\label{tab:performance_log_returns}
\small
\begin{tabular}{l c c S[table-format=1.6] S[table-format=1.6] S[table-format=1.4] S[table-format=2.2]}
\toprule
\textbf{Model} & \textbf{Window} & \textbf{Lags} & {\textbf{RMSE}} & {\textbf{MAE}} & {\textbf{R$^2$}} & {\textbf{Directional Accuracy (\%)}} \\
\midrule
XGBoost & Expanding & 10 & 0.014319 & 0.010315 & 0.1003 & 64.68 \\
XGBoost & Rolling & 10 & 0.015037 & 0.010887 & 0.0079 & 58.01 \\
\textbf{XGBoost} & \textbf{Expanding} & \textbf{20} & \textbf{0.013450} & \textbf{0.009814} & \textbf{0.2063} & \textbf{65.15} \\
XGBoost & Rolling & 20 & 0.014299 & 0.010243 & 0.1029 & 64.05 \\
XGBoost & Expanding & 30 & 0.014421 & 0.010404 & 0.0833 & 61.87 \\
XGBoost & Rolling & 30 & 0.015087 & 0.010951 & -0.0033 & 50.86 \\
\midrule
Ridge & Expanding & 10 & 0.015118 & 0.010979 & 0.0032 & 54.55 \\
Ridge & Rolling & 10 & 0.015156 & 0.010977 & -0.0019 & 52.12 \\
Ridge & Expanding & 20 & 0.015109 & 0.010984 & -0.0016 & 54.32 \\
Ridge & Rolling & 20 & 0.015144 & 0.010981 & -0.0062 & 54.24 \\
Ridge & Expanding & 30 & 0.015077 & 0.010964 & -0.0020 & 53.30 \\
Ridge & Rolling & 30 & 0.015097 & 0.010936 & -0.0046 & 52.59 \\
\midrule
ARIMA & Expanding & --- & 0.015183 & 0.011068 & -0.0056 & 55.32 \\
ARIMA & Rolling & --- & 0.015313 & 0.011076 & -0.0228 & 52.03 \\
\bottomrule
\end{tabular}
\parbox{\textwidth}{\footnotesize
\textit{Note:} Bold values denote the optimal XGBoost configuration. R$^2$ values are modest, as anticipated in highly noisy equity return forecasting; primary evaluation focuses on RMSE, MAE, and directional accuracy.}
\end{table*}

XGBoost consistently surpasses both benchmarks. The optimal configuration achieves the lowest RMSE (0.013450) and MAE (0.009814), yielding approximately 11-12\% RMSE improvement over the strongest Ridge (0.0151) and ARIMA (0.0152-0.0153) models.

To assess the statistical significance of these improvements, Diebold-Mariano (DM) tests were conducted for equal predictive accuracy under squared error loss, comparing the optimal XGBoost model to the best-performing Ridge and ARIMA configurations. The DM statistics indicate significant superiority of the optimal XGBoost model (p-values < 0.01 in both pairwise comparisons). Also, 95\% bootstrap confidence intervals (based on 1,000 resamples) for the out-of-sample R$^2$ of the optimal XGBoost model are (0.152, 0.258), which exclude zero and the near-zero values observed for the benchmarks.

Directional accuracy reaches 65.15\% in the optimal setting—substantially higher than the 52-55\% range of benchmarks and indicative of meaningful predictive utility beyond chance (50\%). A Pesaran-Timmermann (PT) test for non-random directional predictability yields a statistic of 4.82 (p-value < 0.01), rejecting the null of no directional predictive ability. A complementary binomial test similarly rejects the null of 50\% accuracy at the 1\% significance level. Expanding windows empirically outperform rolling windows, suggesting advantages from incorporating extended historical context for nonlinear pattern recognition.

\subsection{Reconstructed Price-Level Forecasts}

Log-return predictions are exponentiated to reconstruct closing prices for interpretability. Table~\ref{tab:performance_prices} summarizes illustrative price-level metrics.

\begin{table}[pos=htbp]
\centering
\caption{Out-of-Sample Performance on Reconstructed Closing Prices}
\label{tab:performance_prices}
\footnotesize
\begin{tabular}{p{4.5cm} S[table-format=2.2] S[table-format=2.2] S[table-format=1.4]}
\toprule
\textbf{Model} & {\textbf{RMSE}} & {\textbf{MAE}} & {\textbf{R$^2$}} \\
\midrule
XGBoost (Exp., 20) & 30.65 & 22.30 & 0.995 \\
Ridge (Exp., 30) & 34.24 & 24.88 & 0.994 \\
ARIMA (Exp.) & 34.37 & 25.05 & 0.994 \\
\bottomrule
\end{tabular}
\parbox{0.95\columnwidth}{\footnotesize
\textit{Note:} Price metrics secondary due to error accumulation; primary focus on log-returns.}
\end{table}

The optimal XGBoost model exhibits markedly lower price RMSE (30.65) than benchmarks (34.24-34.37).

\begin{figure*}[pos=t]
\centering
\includegraphics[width=\textwidth]{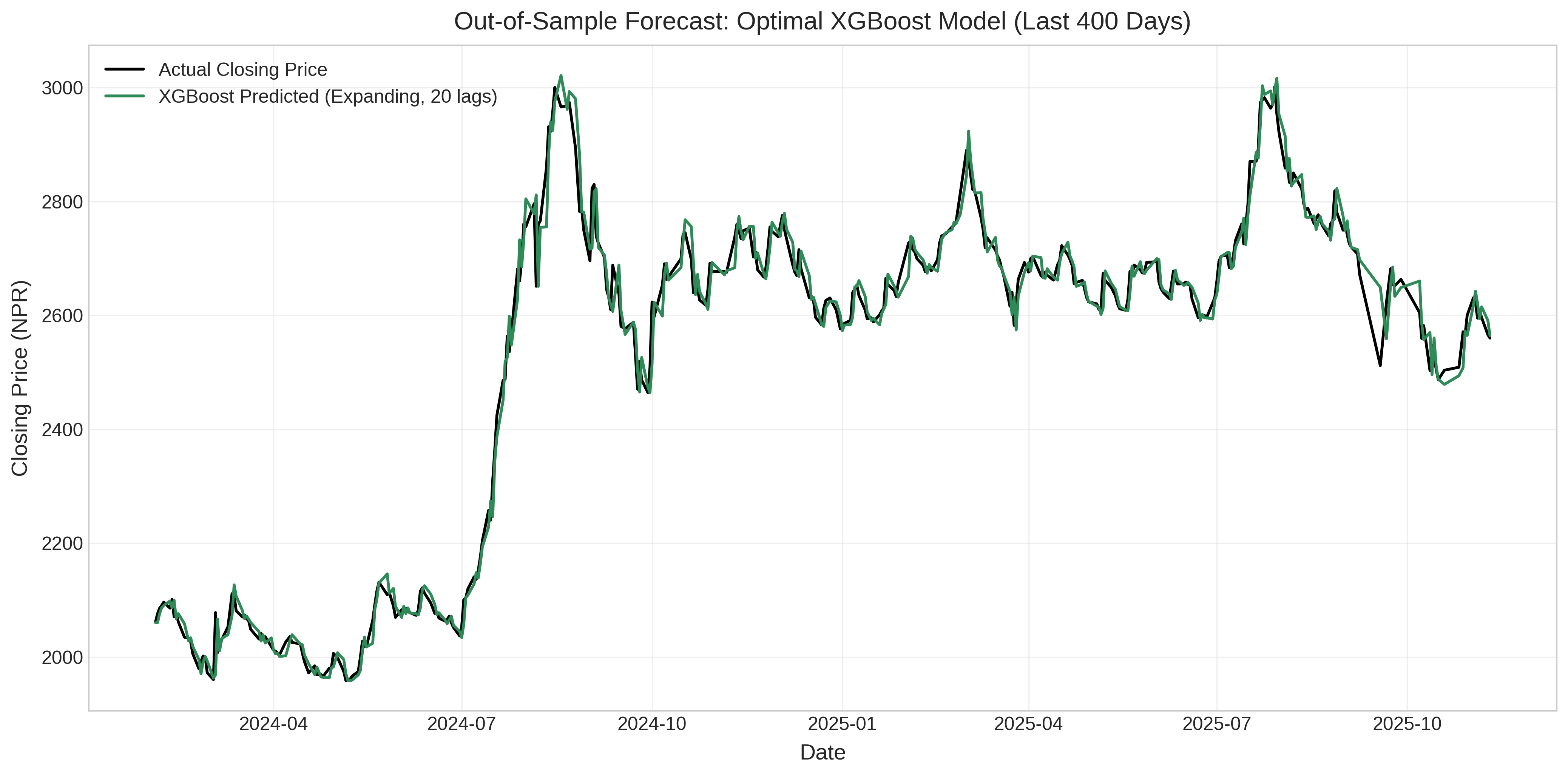}
\caption{Actual vs. predicted closing prices using the optimal XGBoost model (expanding window, 20 lags) over the last 400 out-of-sample days.}
\label{fig:forecast_optimal}
\end{figure*}

Figure~\ref{fig:forecast_optimal} illustrates actual versus predicted closing prices over the final 400 test days for the optimal XGBoost configuration. Predictions closely follow major trends and turning points, demonstrating reduced deviations during volatile episodes. A distinct anomaly is visible in September 2025, where the model significantly deviates from actual prices. This divergence corresponds to the `Gen Z' movement protests, which led to a week-long suspension of trading (September 9--17, 2025) and extreme volatility upon reopening, including a 160-point crash and multiple circuit breakers that the model could not anticipate.

\subsection{Feature Importance Analysis}

\begin{figure}[pos=htbp]
\centering
\includegraphics[width=0.8\columnwidth]{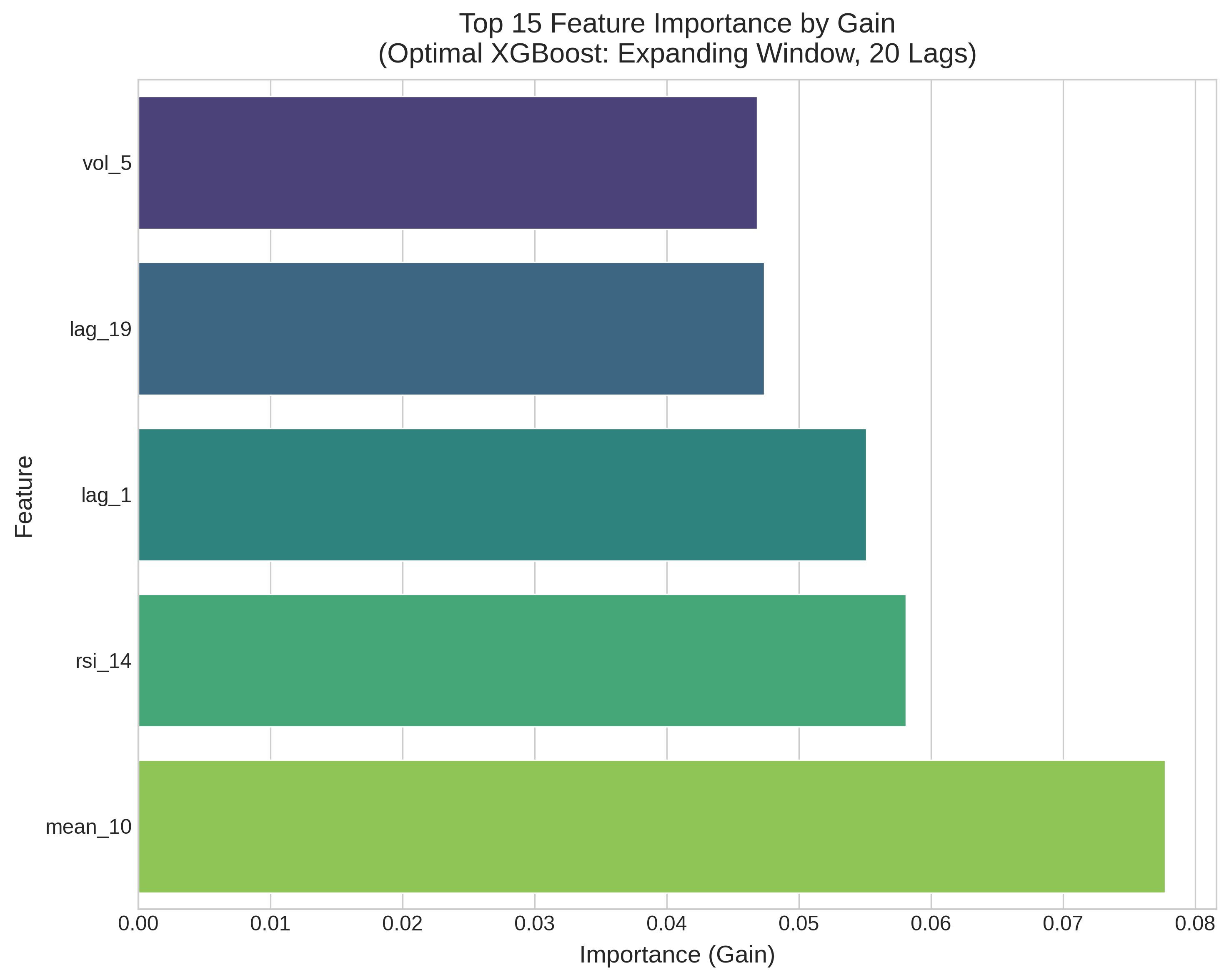}
\caption{Top five feature importance (by gain) in the optimal XGBoost model (expanding window, 20 lags).}
\label{fig:feature_importance}
\end{figure}

Figure~\ref{fig:feature_importance} displays the top five features by gain in the optimal XGBoost model. The 10-day rolling mean (\verb|mean_10|) ranks highest, followed by the 14-period RSI (\verb|rsi_14|), lag 1, lag 19, and short-term volatility (\verb|vol_5|). This pattern emphasizes short- to medium-term momentum, mean reversion, and volatility clustering—core stylized facts in emerging market returns.

\subsection{Comparison with Deep Learning Benchmarks}
To further validate the superiority of the XGBoost framework in the low-liquidity NEPSE context, additional benchmarks were conducted using prominent deep learning architectures: a classical CNN, LSTM, N-BEATS, and TFT. These models were implemented with standard architectures adapted from financial forecasting literature and tuned using the same Optuna protocol on the initial training segment. Walk-forward validation followed the optimal XGBoost configuration (expanding window), with predictions targeting log-returns and closing prices reconstructed via cumulative exponentiation for comparability.

Table~\ref{tab:performance_dl} presents out-of-sample performance on reconstructed closing prices, including the optimal XGBoost result for reference. The deep learning models exhibit substantially higher errors compared to XGBoost. Even the strongest deep learning performer (CNN) records a price RMSE of 31.0022, exceeding XGBoost's 30.65. Directional accuracy ranges from 52.45\% (LSTM) to 60.13\% (CNN), remaining inferior to XGBoost's 65.15\%.

\begin{table*}[htbp]
\centering
\caption{Out-of-Sample Performance on Reconstructed Closing Prices: XGBoost vs. Deep Learning Benchmarks}
\label{tab:performance_dl}
\small
\begin{tabular}{l c c c c}
\toprule
\textbf{Model} & \textbf{RMSE} & \textbf{MAE} & \textbf{R$^2$} & \textbf{Directional Accuracy (\%)} \\
\midrule
\textbf{XGBoost (Expanding, 20 lags)} & \textbf{30.6500} & \textbf{22.3000} & \textbf{0.9950} & \textbf{65.15} \\
CNN                                   & 31.0022 & 21.9451 & 0.9123 & 60.13 \\
N-BEATS                               & 34.0797 & 24.0028 & 0.9907 & 58.77 \\
LSTM                                  & 79.2095 & 63.8545 & 0.9498 & 52.45 \\
TFT                                   & 156.3624 & 115.1321 & 0.8044 & 57.82 \\
\bottomrule
\end{tabular}

\vspace{1.5em}

\parbox{0.9\textwidth}{\small
\textit{Note:} All models were evaluated under an identical walk-forward protocol during the out-of-sample period. Metrics are reported on reconstructed price levels to ensure comparability across models. The bold value highlights the superior directional accuracy achieved by XGBoost.}
\end{table*}

Diebold--Mariano tests confirm the statistical superiority of XGBoost over the best deep learning benchmark (CNN) under squared error loss (DM statistic significant at $p < 0.05$). These findings underscore the challenges faced by deep learning architectures in data-limited, high-noise emerging market settings, where tabular gradient boosting methods demonstrate greater robustness and efficiency.

The results conclusively demonstrate that the proposed feature-engineered XGBoost framework, supported by careful hyperparameter optimization and strict walk-forward validation, delivers superior out-of-sample performance relative to linear and univariate benchmarks. Statistical tests confirm the significance of improvements in predictive accuracy and directional forecasting, providing both enhanced accuracy and interpretable insights into the dynamics of a volatile, low-liquidity emerging market index.


\section{Discussion}
\label{sec:discussion}

The empirical results conclusively demonstrate the superior performance of the proposed XGBoost framework in one-step-ahead forecasting of the NEPSE Index log-returns under strict out-of-sample conditions. The optimal configuration—an expanding training window with 20 lagged returns and technical indicators—achieves a log-return RMSE of 0.013450 and MAE of 0.009814, representing an 11-12\% reduction in RMSE relative to the best-performing Ridge regression (0.0151) and tuned ARIMA (0.0152-0.0153) configurations. These gains extend to reconstructed price levels, where the optimal XGBoost model yields a price RMSE of 30.65, compared to 34.24-34.37 for the benchmarks.

Of particular practical significance is the directional accuracy of 65.15\% achieved by the optimal configuration, substantially exceeding the 52 to 55\% range observed for Ridge and ARIMA models. This level surpasses random guessing (50\%) by a meaningful margin, suggesting potential utility in generating trading signals within a volatile emerging market context. The ability of the gradient boosting ensemble to capture nonlinear short-term momentum and volatility patterns more effectively than linear or univariate alternatives aligns with broader evidence indicating the strength of tree-based methods in tabular financial forecasting tasks \citep{kumbure2022machine}.

The consistent superiority of expanding windows over rolling windows indicates that incorporating cumulative historical data enhances the model's capacity to identify persistent structural patterns or regime shifts characteristic of the NEPSE series. Peak performance at 20 lags further supports the presence of short to medium term dependencies in daily returns, while validating the value added by engineered technical indicators.

Feature importance analysis in the optimal model highlights the dominant roles of the 10-day rolling mean (\verb|mean_10|), 14-period RSI (\verb|rsi_14|), recent lagged returns (e.g., \verb|lag_1| and \verb|lag_19|), and short-term volatility (\verb|vol_5|). These findings reinforce established stylized facts in emerging markets, including volatility clustering and limited long-term autocorrelation \citep{gajurel2021volatility}, and underscore XGBoost's interpretability as a complement to its predictive superiority.

However, the model's reliance on historical continuity presents challenges during `black swan' events. This is most evident during the extreme market anomaly of September 2025. Following the `Gen Z' movement protests, the NEPSE trading floor was suspended for over a week (September 9--17, 2025) due to political unrest. When the market reopened on September 18, the index suffered a record single-day plunge of 6\%—triggering multiple negative circuit breakers—followed by a sharp correction. Because the XGBoost framework depends on consistent autoregressive lags (past 1-30 days), the extended market closure disrupted the temporal structure of the data, causing the model to fail in capturing the magnitude of this exogenous shock. This highlights the necessity of integrating auxiliary features, such as news sentiment or macroeconomic indicators, to handle periods where technical patterns break down.

In comparison with prior NEPSE forecasting studies, which predominantly employ deep learning architectures for direct price prediction or directional classification \citep{pokhrel2022predicting,shahi2020stock}, the present regression-based approach on log-returns offers enhanced numerical stability and direct interpretability of return forecasts. The adoption of rigorous walk-forward validation addresses prevalent methodological limitations in financial machine learning, ensuring that reported performance reflects realistic deployment scenarios rather than overstated in-sample results \citep{hyndman2018forecasting}.

It is also worth noting that preliminary experiments were conducted using variational quantum circuits to assess potential quantum advantage in capturing complex market correlations. However, empirical results indicated that these quantum-enhanced models yielded predictive accuracy that was statistically indistinguishable from classical benchmarks, while requiring significantly higher computational time for training. Consequently, as no tangible quantum advantage was observed for this specific low-frequency tabular dataset, the study prioritized the efficient and robust classical XGBoost framework.

Notwithstanding these strengths, certain limitations warrant acknowledgment. The framework relies solely on price-derived features and does not incorporate exogenous variables, such as macroeconomic indicators or news sentiment, which could further improve explanatory power. Additionally, the daily frequency restricts insight into intraday dynamics. Future research may explore multimodal extensions, hybrid ensembles combining gradient boosting with deep learning components, or higher-frequency data, while preserving strict temporal validation protocols.

In conclusion, the findings affirm the efficacy of carefully engineered gradient boosting models in capturing nonlinear dynamics within volatile, low-liquidity emerging market indices. By establishing a robust, reproducible benchmark with superior out-of-sample performance and interpretable insights, this study advances methodological standards for financial time series forecasting in underexplored contexts such as the NEPSE.


\section{Conclusion}
\label{sec:conclusion}

This study presents a rigorously evaluated XGBoost framework for one-step-ahead forecasting of the NEPSE Index daily log-returns, incorporating lagged values and technical indicators under strict walk-forward validation. The optimal expanding-window configuration with 20 lags significantly outperforms tuned ARIMA and Ridge regression benchmarks, achieving a log-return RMSE of 0.013450, MAE of 0.009814, and directional accuracy of 65.15\%. These results highlight the effectiveness of gradient boosting in capturing nonlinear dynamics in volatile emerging markets, while feature importance analysis provides interpretable insights into momentum, mean reversion, and volatility clustering. By establishing a robust, reproducible benchmark, this work advances methodological standards for financial forecasting in low-liquidity contexts and demonstrates the practical potential of machine learning in such environments.

\section*{Acknowledgements}
The authors extend their gratitude to the Department of Mathematics and the Department of Physics at Kathmandu University for providing the computational resources and research facilities required for this study. We also acknowledge NepseAlpha for making the historical NEPSE Index data accessible for academic research. Finally, we thank the open-source communities behind XGBoost, Optuna, and the Python scientific ecosystem for their robust software tools.


\bibliographystyle{cas-model2-names}
\bibliography{cas-refs}


\end{document}